\documentclass[journal]{vgtc}                     

\onlineid{0}

\vgtccategory{Research}

\title{SPHERE: Adaptive VR Indoor Scene Generation via LLM-Enhanced Spatial Preference Learning and Human-in-the-Loop RL}

\author{%
\authororcid{Hyeonmin Lee}{0009-0004-1153-1023},
\authororcid{Zheng Wei}{0000-0001-7444-2547},
\authororcid{Kyungmin Kwon}{0009-0001-3082-5172},
\authororcid{Jumin Seo}{0009-0002-8393-5428},
\authororcid{Jiwon Park}{0009-0007-8192-617X},
\authororcid{Hayoung Oh}{0000-0002-7362-5138}\textsuperscript{*},
}

\authorfooter{
    \item
    Hyeonmin Lee, Kyungmin Kwon, Jumin Seo, Jiwon Park, and Hayoung Oh
    are with Sungkyunkwan University.
    E-mail: llhyeonmin12@gmail.com, km0228kr@skku.edu,
    zox2mm@g.skku.edu, orilove1@g.skku.edu, and hyoh79@skku.edu.

    \item
    Zheng Wei is with The Hong Kong University of Science and Technology.
    E-mail: zwei302@connect.ust.hk.
}

\teaser{
  \centering
  \includegraphics[width=\linewidth]{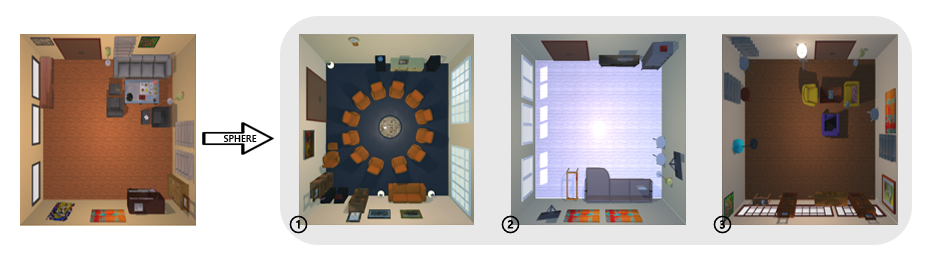}
  \caption{SPHERE incrementally adapts subsequent VR indoor scenes by inferring user preferences and spatial tendencies from an initial user-edited scene (left). Across the generated environments (1-3), SPHERE reflects user-specific layout patterns, stylistic and design preferences, as well as object-level tendencies.}
  \label{fig:teaser}
}

\abstract{%
While Large Language Models (LLMs) advance 3D indoor scene synthesis, current pipelines fail to retain user-specific preferences across sessions, making immersive authoring a repetitive and physically fatiguing process. We present SPHERE, an adaptive VR generation framework that transforms isolated synthesis into continuous human-AI co-creation. SPHERE extracts persistent spatial preferences from natural multimodal interactions (speech and controller edits). To ensure geometric resilience against spatial distortions, it abstracts these raw edits into hierarchical constraints modeling both local functional and global topological contexts. Furthermore, a human-in-the-loop reinforcement learning mechanism dynamically updates retrieval policies based on the user’s final edited scenes. A mixed-design user study ($N=42$) and an offline ablation demonstrate that SPHERE significantly reduces corrective edits and physical demand, preventing bias toward shallow object-level traits to yield geometrically resilient, profile-aligned layouts. Ultimately, SPHERE demonstrates how capturing demonstrated spatial logic enables controlled spatial adaptation, establishing a reliable, governed human-AI collaboration framework for immersive authoring. Project page and source code will be available at:
\url{https://github.com/hyeonmin11/SPHERE}}


\graphicspath{{figs/}{figures/}{pictures/}{images/}{./}} 

\usepackage{multirow}
\usepackage{booktabs}                  
\usepackage{lipsum}                    
\usepackage{mwe}                       
\usepackage{ccicons}                   

\usepackage{mathptmx}                  
\usepackage{amsmath}   
\begin{document}
\providecommand{\manuscriptnotetxt}{}

\firstsection{Introduction}

\maketitle

Recent advancements in 3D indoor scene synthesis (3DISS) have increasingly leveraged powerful generative models to construct complex, object-centric environments. Contemporary systems equipped with Large Language Models (LLMs) and multimodal architectures can now dynamically encode user-provided text to guide zero-shot object selection and spatial placement \cite{birr2024autogpt+, huang2025fireplace, li2024llm, honerkamp2024language, tang2023cotdet}. However, immersive spatial design is not merely a one-off generation task; it is an ongoing process of centering humans to shape intelligence. In this process, establishing a governed AI collaboration framework is essential to reliably bridge human intent with predictable AI outcomes. Despite algorithmic leaps in text-to-3D generation, end-to-end LLM-driven pipelines exhibit a systemic shortcoming: they systematically fail to explicitly model the latent evolution of user preferences across repeated, temporally disjoint interactions. Existing sequential editing frameworks \cite{hou2025echoladder, wang2025can, sun20253d} predominantly treat user modifications as isolated, context-free error corrections. Consequently, users are burdened with repeatedly executing identical corrective operations across discrete authoring tasks, which significantly increases interaction friction and physical fatigue in VR environments. Furthermore, prevailing spatial reasoning mechanisms often anchor spatial relationships to rigid absolute Cartesian coordinates \cite{gao2023scenehgn, li2024llm} or flat, unstructured semantic graphs \cite{honerkamp2024language, liu2025semantics}. These formulations inherently skew heavily toward extracting simple, low-level object traits and frequently lead to catastrophic layout failures when the generated spatial volume undergoes geometric scaling or aspect-ratio distortion. To overcome these critical limitations and advance the paradigm of human-AI co-creation, we introduce \textbf{SPHERE}, a novel LLM-enhanced spatial preference modeling framework with human-in-the-loop reinforcement learning for adaptive VR indoor scene generation. SPHERE transforms one-off scene synthesis into a continuous, personalized co-creation process by introducing an explicit, persistent preference memory architecture. Instead of relying on cognitively demanding explicit ratings, SPHERE innovatively captures natural, multimodal editing behaviors in VR—specifically fusing continuous speech-driven instructions with controller-grounded spatial manipulations—as continuous implicit behavioral signals. To ensure geometric resilience, the system abstracts these raw interactions into hierarchical contextual constraints by algorithmically inferring both localized functional relationships (Local Context) and macro-level global topological rules (Global Context). To dynamically apply these customized spatial constraints to the generative engine for future sessions, SPHERE utilizes a human-in-the-loop policy optimization mechanism. By calculating a deterministic reward based on whether the sampled functional affordances were successfully materialized in the user's terminal scene state, the system continuously updates a cross-attention reranker via a policy-gradient objective. This reinforcement loop ensures that synthesized layouts inherently conform to the user's previously demonstrated latent logic, facilitating robust zero-shot personalization without necessitating computationally expensive model fine-tuning. 

In summary, the primary contributions of this work are three-fold: 

1) \textbf{Adaptive VR Framework:} We propose SPHERE, which integrates persistent preference memory and human-in-the-loop RL to minimize interaction friction and repetitive editing in immersive environments. 

2) \textbf{Context-Aware Spatial Reasoning:} We extract hierarchical spatial constraints (object, intra-area, and inter-area) from multimodal inputs, ensuring that learned spatial rules remain topologically stable across varying room scales and aspect ratios. 

3) \textbf{Empirical Validation:} Through a controlled user study ($N=42$) and an offline ablation, we demonstrate that SPHERE significantly reduces corrective edits and physical effort while producing layouts that robustly align with target spatial profiles without requiring full-model fine-tuning.


\section{Related Work}

\subsection{Language-Driven 3D Scene Generation}
Recent advancements in 3D indoor scene synthesis (3DISS) have increasingly leveraged powerful generative models to construct complex, object-centric indoor environments \cite{gao2023scenehgn, wang2021sceneformer, kim2026scenelinker, yin2024text2vrscene}. Early computational approaches within this domain heavily relied on conventional natural language processing (NLP) techniques to map user instructions to curated 3D assets \cite{chang2017sceneseer, ma2018language}. These sequential pipelines typically functioned by constructing explicit structural scene graphs from textual inputs, aligning them with pre-existing database topologies, and subsequently enriching partially matched subgraphs with supplementary object instantiations. However, the practical scalability and creative utility of these heuristic methods were intrinsically bottlenecked by limited object category taxonomies and the restricted geometric diversity of the underlying datasets. The advent of Large Language Models (LLMs) \cite{brown2020language} and advanced multimodal architectures has fundamentally shifted this paradigm. Contemporary 3DISS systems can now dynamically encode free-form, user-provided textual input into high-dimensional vector representations, effectively guiding zero-shot object selection and spatial placement \cite{liu2023clip, chen2020intelligent}. Despite these algorithmic leaps, end-to-end LLM-driven generative pipelines exhibit notable systemic shortcomings. Although recent frameworks have largely mitigated spatial implausibilities and intersecting bounding volumes through physically grounded refinement mechanisms, they often over-optimize for geometric non-collision\cite{ccelen2024design, yang2024llplace}. Consequently, these systems frequently neglect common-sense reasoning regarding object functionality and human accessibility, yielding placements that are physically valid but practically unusable. Furthermore, while recent frameworks have introduced sequential editing to support iterative scene refinement \cite{hou2025echoladder, wang2025can, huang2025fireplace, abdelreheem2025placeit3d}, they systematically fail to explicitly model the latent evolution of user preferences across repeated, temporally disjoint interactions. Consequently, users are burdened with repeatedly executing identical corrective operations across discrete authoring tasks, significantly increasing interaction friction. To overcome this critical limitation, our work introduces a novel mechanism designed to bridge the semantic gap between isolated generation instances. By proposing an explicit, persistent preference memory architecture, we successfully transform one-off text-to-3D generation into a continuous, personalized human-AI co-creation process.


\subsection{Preference Learning from User Interaction}
The paradigm of aligning algorithmic outputs with human intent has been extensively formalized through the lens of Reinforcement Learning from Human Feedback (RLHF) \cite{ziegler2019fine, ouyang2022training}. Moving beyond simplistic pairwise preference rankings, contemporary alignment research has increasingly utilized free-form textual feedback \cite{fernandes2023bridging} to dynamically rank candidate outputs and iteratively fine-tune generative models to maximize preferred semantic refinements. In spatially grounded and embodied domains, such as robotic navigation\cite{kant2022housekeep, cheng2023llf, aroca2024predict}, hindsight behavioral feedback has also been successfully leveraged to supervise policy learning and correct navigational trajectories. 

As LLM-based autonomous agents are rapidly deployed in real-world interactive applications, robust personalization has emerged as a paramount computational challenge. While techniques such as Supervised Fine-Tuning (SFT) \cite{xie2024monte, zhang2024rest}, online Direct Preference Optimization (DPO) \cite{guo2024direct}, and continuous reinforcement learning \cite{zheng2026lifelong, wang2025ragen} can theoretically improve user alignment, they are often computationally prohibitive, require extensive data curation, and are exceedingly difficult to scale dynamically across diverse multi-user deployments. Consequently, computationally lightweight, prompt-centric personalization paradigms—encompassing personalized context retrieval \cite{gao2024aligning, tan2025prospect}, dynamically learned prompt policies \cite{liu2023llm, al2024can, shinn2023reflexion, zhao2024expel}, and algorithmic prompt rewriting techniques \cite{kong2024prewrite, sarkar2025conversational}—have garnered substantial architectural attention. A parallel line of research investigates output revision and edit-intent modeling, which historically targets deterministic error correction, such as source code repair \cite{reid2022diffuser, chen2019sequencer} or factual hallucination refinement \cite{balachandran2022correcting}. Methodologies in this space include edit-intent modeling \cite{kim2022improving, liu2022second, chen2025analyzing}, latent edit representations\cite{marrese2023edit, gao2024aligning, yin2018learning}, structured hierarchical trees \cite{zhou2310language, yao2021learning}, and the modeling of atomic edit sequences \cite{laban2023swipe, ma2016action}. However, these approaches predominantly treat edits as isolated, context-free corrections and do not model the overarching evolution of individual user preferences over repeated interactions. Unlike traditional methods that require cognitively demanding explicit human ratings or computationally heavy model parameter updates, our research innovatively investigates how natural editing behaviors in immersive environments can serve as continuous, implicit behavioral signals. This enables the incremental alignment of automated scene generation with highly individualized spatial logic without interrupting the user's creative flow.

\subsection{Context-Aware Spatial Reasoning in Immersive Environments}
A foundational challenge within interactive scene authoring is the semantic translation of discrete, low-level user manipulations into coherent, high-level spatial intelligence. To impose structural order on generated environments, some approaches \cite{fu2024anyhome, deitke2022️} have adopted Semantic Asset Group (SAG) concepts to mathematically identify commonly co-occurring object clusters within a scene graph. Furthermore, constraint-based and heuristic-driven scene synthesis engines have long explored sophisticated mechanisms to inject commonsense spatial reasoning and physical priors into automated scene planning \cite{yang2024holodeck, chang2014interactive}. Nevertheless, the prevailing literature predominantly anchors spatial relationships to rigid absolute coordinates \cite{aguina2024open, choi2026scenenat} or flat, unstructured semantic graphs \cite{hughes2024foundations, purkait2020sg, para2021generative, wei2023lego}. These traditional formulations are inherently biased; they frequently skew heavily toward extracting simple, low-level object traits rather than capturing the complex structural intent behind a user's bespoke spatial arrangement. Relying on absolute Cartesian framing leads to catastrophic layout failures when the generated spatial volume undergoes geometric scaling or aspect-ratio distortion. Our work explicitly addresses this critical need for deeper, more resilient spatial intelligence by abstracting raw spatial interactions into hierarchical contextual constraints. By algorithmically inferring both localized functional relationships and macro-level global topological rules, we ensure that the system's learned preferences are geometrically adaptable. This hierarchical representation allows the extracted user constraints to remain robust and highly transferable, successfully governing AI synthesis even under significant spatial variations, such as dramatic changes in room scale and structural aspect ratios.

\begin{figure}[t]
    \centering
    \includegraphics[width=\linewidth]{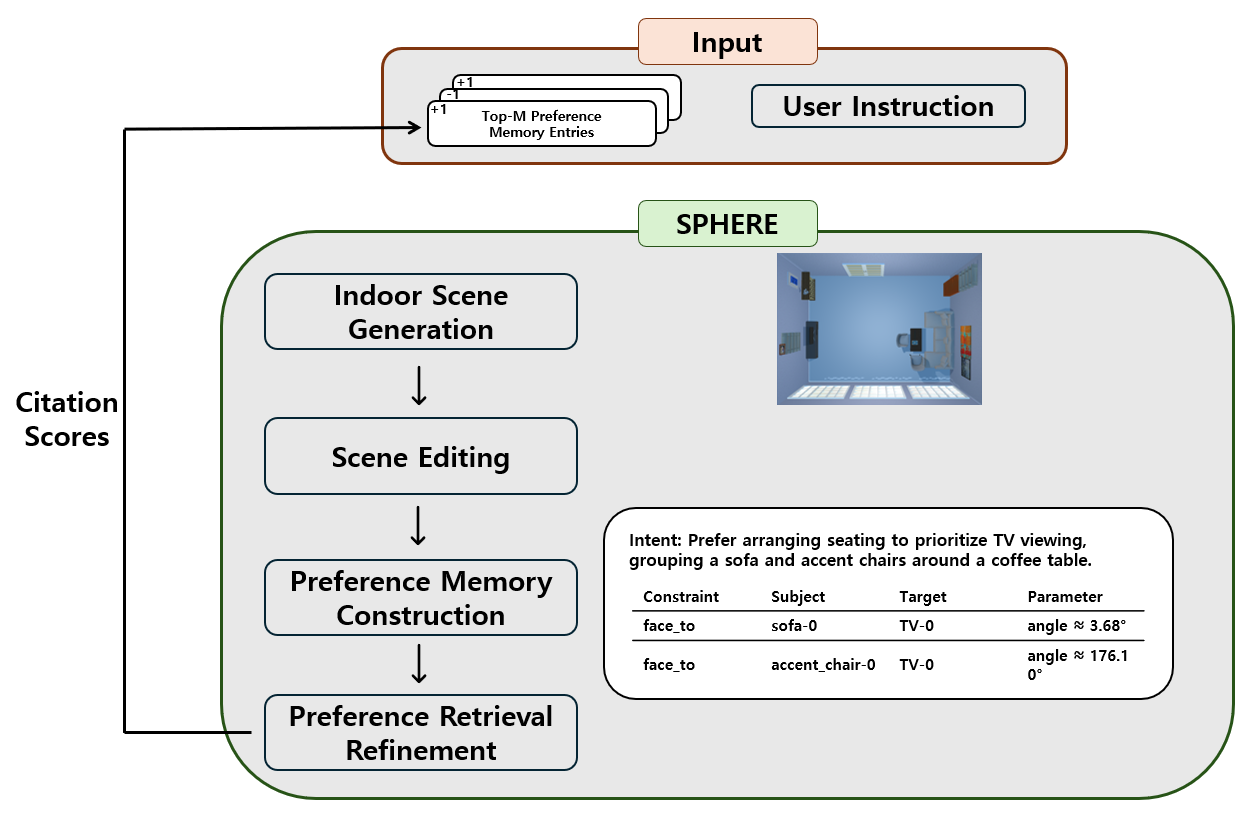}
    \caption{Given user instruction and the Top-M retrieved preference memory entries as input, SPHERE performs indoor scene generation to produce an initial layout. The generated scene is then refined through scene editing, where users interactively adjust indoor scene in VR. From the edited scene, preference memory construction extracts object-level and layout-level constraints from the edited scene. Finally preference retrieval refinement updates rewards using realization signal, closing the feedback loop and enabling incremental personalization across sessions.}
    \label{overview}
\end{figure}

\section{Methods}

In this section, we delineate the architecture of SPHERE, an adaptive VR framework designed for indoor scene personalization. SPHERE systematically captures latent user preferences from interactive editing sessions, formalizing these behavioral signals into reusable spatial constraints for subsequent generative tasks. The core generative foundation of SPHERE relies on the Holodeck engine \cite{yang2024holodeck}, which facilitates room-aware object placement and constraint-driven synthesis. To ensure large-scale, semantically diverse scene compositions, the framework leverages the Objaverse 3D asset dataset \cite{deitke2023objaverse}. Since Holodeck was originally implemented as a single-turn 3D scene generation system that does not account for either user authoring or VR, we modified it to support authoring in an immersive environment by (1) enabling bidirectional synchronization with the Unity/VR runtime and (2) adjusting the properties of scene elements so that users can explore and interact with objects in VR. While foundational generative engines like Holodeck excel at zero-shot scene synthesis, they inherently lack temporal persistence regarding user-specific spatial tendencies. SPHERE bridges this gap by introducing an explicit, persistent preference memory mechanism. As illustrated in Figure~\ref{overview}, the operational pipeline encompasses four sequential stages. Initially, the system synthesizes a baseline indoor environment using Holodeck, conditioned on user instructions and top-ranked historical preference memory entries. Second, users are immersed in the VR environment, where they iteratively refine the scene via multimodal inputs, specifically fusing natural language utterances with controller-grounded spatial manipulations. Third, upon session termination, SPHERE extracts and structures preference memory entries from both the terminal scene geometry and the comprehensive interaction log. Finally, an online reinforcement learning objective reranks retrieval weights based on the realized affordances, dynamically injecting these customized spatial constraints into the generative engine for future sessions. 

\subsection{Multimodal Scene Editing}

\subsubsection{Multimodal Fusion and Intent Resolution}
To support fluid authoring, the VR interface fuses locomotive exploration, object selection, and continuous speech-driven instruction. Users navigate the virtual space via continuous joystick locomotion, an exploratory phase that operates independently of the system's semantic interpretation layer. Authoring is initiated via controller input, capturing unconstrained, free-form natural language utterances, which are transcribed in real-time utilizing the Whisper ASR model. For spatially grounded commands, users employ the controller trigger for object selection and the grip button to designate geometric target coordinates. The chronological sequence of these selections is strictly preserved to maintain the temporal structure of the user's explicit intent. These heterogeneous inputs are fused into a structured interaction state formulation:
\begin{verbatim}
STATE = { "instruction": "<raw utterance>",
    "objects": [obj_1, obj_2, ...],
    "positions": [pos_1, pos_2, ...]}
\end{verbatim}

This state is transmitted to a Large Language Model (LLM)-driven intent resolver. The resolver performs coreference resolution, mapping linguistic variables to their corresponding spatial entities or coordinates, thereby generating a fully deterministic and explicit instruction.

\begin{figure}[t]
    \centering
    \includegraphics[width=\linewidth]{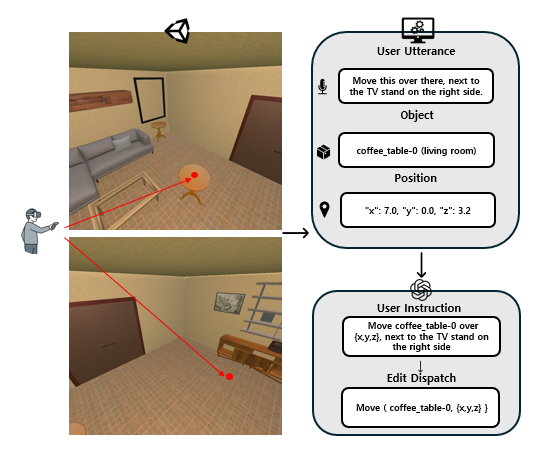}
    \caption{Users interactively modify objects within the VR environment through natural language utterances and controller-based interactions that specify target objects and spatial positions. The system parses the multimodal input to identify the referenced object and desired placement, converts it into a structured instruction, and dispatches the corresponding edit operation to update the scene.}
    \label{fig:interface}
\end{figure}

\subsubsection{Deterministic Edit Execution}
The resolved instruction undergoes semantic decomposition via the LLM, translating the high-level directive into an ordered sequence of atomic scene-editing subtasks (see Figure~\ref{fig:interface}). Each subtask represents a singular, actionable primitive (e.g., translation, rotation, instantiation). The system enforces strict mapping constraints to ensure subtask serialization strictly mirrors the user's implied procedural intent. These subtasks are dispatched to a predefined set of scene operators: Add, Move, Rotate, Scale, Light adjustment, and Remove. For the \textit{Add} operator, the system executes a semantic asset retrieval pipeline. SPHERE retrieves and instantiates the highest-scoring candidate asset based on CLIP similarity\cite{radford2021learning} between the natural-language subtask query and GPT-4-V-generated annotations of Objaverse assets already implemented in HOLODECK. Operators are applied synchronously to the underlying scene graph (maintained as a JSON schema), and the VR viewport is updated post-execution to reflect the terminal state.


\subsection{Hierarchical Preference Memory Representation}


We deliberately instantiate preference memory post-session rather than via continuous real-time (intra-session) updates, motivated by three key technical and HCI considerations. First, regarding {cognitive load and creative flow}, continuously altering spatial topology or layout heuristics mid-session would introduce visual and structural churn, severely disrupting the user's immersive authoring flow and sense of agency. Second, regarding {computational latency}, extracting hierarchical spatial contexts via DBSCAN and running LLM-based spatial reasoning incur inference overheads that would inevitably degrade high-frame rate VR interactivity if executed synchronously. Third, regarding {intent stability}, immersive scene editing is an inherently iterative process characterized by frequent trial-and-error and temporary object placements. Therefore, evaluating the terminal scene state ensures that the extracted spatial preferences reliably reflect the user's finalized, deliberate layout logic rather than transient, exploratory actions. The memory payload aggregates the terminal scene configuration, the complete multimodal interaction history, and hierarchically derived scene contexts.

\begin{figure}[t]
    \centering   \includegraphics[width=\linewidth,height=0.8\columnwidth,keepaspectratio]{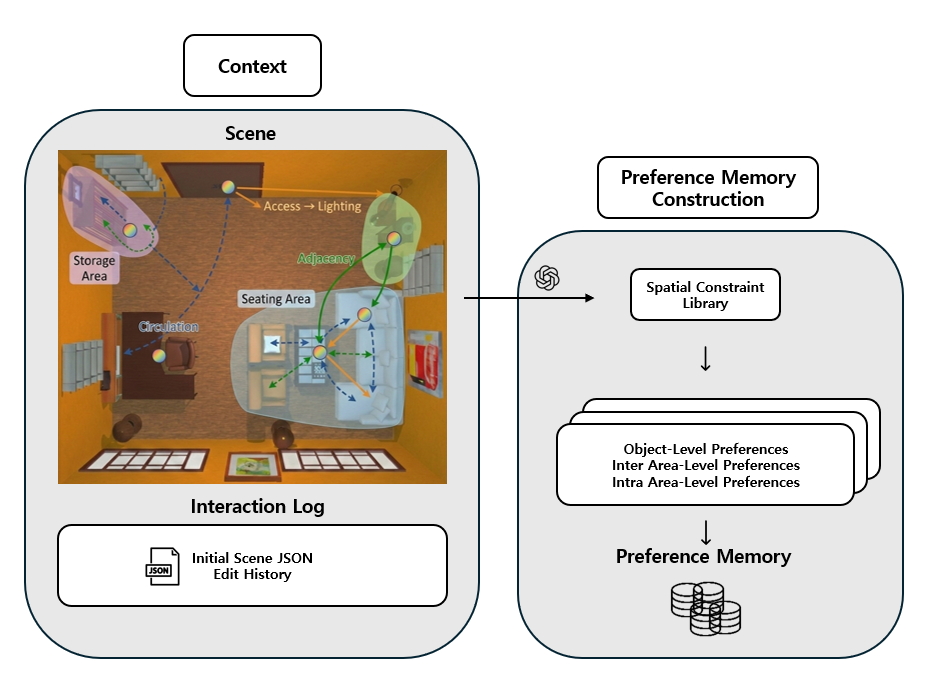}
    \caption{From the scene context and user interaction logs, SPHERE constructs structured preference representations using a spatial constraint library. The extracted preferences are organized hierarchically into object-level, intra-area-level, and inter-area-level constraints, and stored in the preference memory.}
    \label{fig:preference}
\end{figure}

\subsubsection{Contextual Abstraction and Affordance Inference}
As depicted in Figure~\ref{fig:preference}, SPHERE derives spatial abstractions at both local and global scales prior to preference inference. 

\textbf{Local Context:} We extract lower-dimensional object features (pose, ID, semantic category) and apply DBSCAN \cite{ester1996density} on $XZ$-coordinates to cluster objects into functional zones. Within each cluster, we derive pairwise geometric relations using rule-based spatial predicates. A \emph{support} relation is assigned when the horizontal projections of two objects’ bounding volumes overlap and the bottom of the upper object’s bounding volume is vertically adjacent to the top of the lower object’s bounding volume; the lower object is designated as the source and the upper object as the supported target. An \emph{alignment} relation is assigned when two objects are nearly and their yaw difference is smaller than the threshold value, capturing similarly oriented objects within the same functional arrangement. The resulting object composition and geometric relations are subsequently provided to an LLM for semantic interpretation. The LLM first assigns each cluster a generic functional-role label, such as \emph{seating area}, \emph{dining area} or \emph{working area}, based on the semantic categories of its constituent objects. It then identifies functional relationships that may not be directly recoverable from geometry alone, such as chair being used with a desk or a lamp providing illumination for a table. Each inferred relationship is accompanied by a concise affordance-based explanation describing the activity enabled by the object pair. The final local context therefore combines explicit geometric constraints with LLM-inferred functional roles and object-level affordances, representing both the physical organization and intended use of each functional area.

\textbf{Global Context:} 
We construct a room-scale representation by abstracting each local functional zone by its functional role, geometric centroid, and spatial extent. The centroid of each zone is computed from the geometric center of the combined ground-plane projections of the bounding volumes of its constituent objects, while bounding volume is estimated from their minimum and maximum $XZ$-coordinates. These geometric descriptors are used to derive pairwise inter-area connectivity. Two functional zones are considered \emph{adjacent} when the minimum distance between their bounding volumes is below a predefined spatial threshold. To estimate \emph{circulation}, we examine the straight-line path between representative points of the two zones against the room boundary and obstacle geometry. A circulation relation is established only when the path remains unobstructed and provides sufficient lateral clearance for human movement. The resulting inter-area connectivity therefore encodes multiple relations for an area pair, including adjacency and circulation. Finally, an LLM jointly analyzes the functional-role labels and the derived spatial connectivity cues to infer higher-level functional dependencies between zones. These directed dependencies capture task- and affordance-oriented relationships that cannot be determined from geometry alone, such as a work area requiring convenient access to storage or a seating area depending on a nearby lighting zone. By combining zone-level geometry, navigable connectivity, and semantic dependencies, the global context represents the macroscopic functional dependencies across the environment.

\subsubsection{Formal Spatial Constraint Extraction}
Based on the constructed memory payload, LLM is prompted to identify a small set of high-level interior design preferences, spanning object-level, intra-area level, and inter-area level. As shown in Figure~\ref{overview}, these preferences are expressed as natural-language affordance descriptions, and each inferred affordance is rigorously mapped to a symbolic spatial constraint utilizing a predefined relational vocabulary:\begin{equation}
\begin{aligned}
r \in \{&\text{near}, \text{far}, \text{left\_of}, \text{right\_of}, \text{in\_front\_of}, \text{behind\_of}, \text{on\_top\_of}, \\
        &\text{face\_to}, \text{centeraligned\_front}, \text{centeraligned\_side}\}
\end{aligned}
\end{equation}


Constraints are formalized as relational triplets $(o_A, r, o_B)$. These symbolic representations are deterministically grounded into explicit geometric parameters. Specifically: \texttt{near} / \texttt{far}: Defined by the Euclidean norm between the spatial centroids of $o_A$ and $o_B$. \texttt{left\_of} / \texttt{right\_of}: Evaluated via the signed horizontal scalar offset $\Delta x = x_A - x_B$. \texttt{in\_front\_of} / \texttt{behind\_of}: Evaluated via the signed depth scalar offset $\Delta z = z_A - z_B$. \texttt{ontop}: Evaluated via the vertical scalar offset $\Delta y = y_A - y_B$. \texttt{centeraligned\_front} / \texttt{centeraligned\_side}/: Evaluated via centroid alignment offsets along the corresponding spatial axis, using $\Delta z$ for front alignment and $\Delta x$ for side alignment.

The resulting preference memory consists of a scene label and a compact list of preference entries, where each entry contains a natural-language affordance description and its associated spatial constraints.

\subsection{Human-in-the-Loop Policy Optimization}

\subsubsection{Dual-Stage Retrieval and Reward Signal Formulation}




SPHERE employs a dual-stage retrieval mechanism inspired by the retrieval design proposed in \cite{tan2025prospect} to let SPHERE dynamically adjust its retrieval strategy to user preferences. First, the system employs GTE \cite{li2023towards} to encode the user query $q$ and scene labels, and retrieves the top-$K$ candidate scenes based on semantic similarity. A cross-attention reranker then processes the top-$K$ retrieved scene label embeddings, refining their relevance to the user query. The adapted query embedding $q'$ and scene label embeddings $m_i'$ are used to compute relevance scores via dot product, $s_i = q'^{\top} m_i'$. The top-$M$ scene entries are then stochastically sampled using Gumbel-Softmax distribution\cite{gumbel1954statistical} with Gumbel noise\cite{maddison2014sampling}. For each selected scene, its associated affordances are processed through the same retrieval-reranker pipeline, except that the similarity is computed between the user query and affordances. Ultimately SPHERE gets top-$N$ affordances. To close the reinforcement loop, SPHERE outputs a reward based on the user's terminal scene state. We prompt the vision-language model with the rendered top-down orthogonal view of the scene to evaluate if the sampled affordances used for scene generation were successfully materialized, as most core functional topological relationships in indoor layouts are grounded in the XZ-plane. We define a discrete realization signal for each affordance $a_i$:
\begin{equation}
    R_i \in \{+1, -1\}
\end{equation}
where $+1$ denotes successful realization and $-1$ denotes failure or user-driven spatial rejection. 

The composite reward function integrates semantic similarity $s_i$ with the realization signal:
\begin{equation}
    R_i^{\text{scene}} = \begin{cases} s_i^{\text{scene}}, & \text{if } R_i = +1 \\ -(1 - s_i^{\text{scene}}), & \text{if } R_i = -1 \end{cases}
\end{equation}

\begin{equation}
    R_i^{\text{aff}} = \begin{cases} s_i^{\text{scene}}\cdot s_i^{\text{aff}}, & \text{if } R_i = +1 \\ -(1 - s_i^{\text{scene}}\cdot s_i^{\text{aff}}), & \text{if } R_i = -1 \end{cases}
\end{equation}
We design the reward to reflect both binary feedback and semantic similarity $s_i$. Negative samples with larger $s_i$ value are treated as near-miss cases and penalized less, whereas those with smaller 
$s_i$ value receive stronger penalties as clear mismatches. This encourages the model to suppress irrelevant candidates while preserving semantically plausible alternatives.

\subsubsection{Policy Gradient Optimization}
The cross-attention rerankers are iteratively updated leveraging a policy-gradient objective. By aggregating the log-probabilities of sampled scene selections ($\log p^{\text{scene}}$) and affordances ($\log p_i^{\text{aff}}$), the loss function for a given affordance $a_i$ is defined as:
\begin{equation}
    L_i = - \left( R_i^{\text{scene}} \cdot \log p^{\text{scene}} + R_i^{\text{aff}} \cdot \log p_i^{\text{aff}} \right)
\end{equation}

The total episodic loss is the summation over all affordances $L = \sum_i L_i$. Gradients are exclusively routed to update the reranker parameters, while the GTE retriever remains frozen to avoid computationally expensive fine-tuning and to prevent catastrophic forgetting. This is implemented through LoRA modules\cite{hu2022lora} attached to the rerankers to enable lightweight policy updates.

During subsequent inference phases, the baseline generative constraints ($C_{\text{room}}$) synthesized by Holodeck are mathematically unioned with the dynamically retrieved, user-specific affordance constraints ($C_{\text{pref}}$):
\begin{equation}
    C_{\text{final}} = C_{\text{room}} \cup C_{\text{pref}}
\end{equation}

Object selection, environmental atmosphere, and geometric placements are systematically biased to satisfy $C_{\text{final}}$. This mechanism ensures that synthesized layouts inherently conform to the user’s previously demonstrated latent logic, facilitating robust zero-shot personalization without necessitating computationally expensive model fine-tuning.

\section{User Study Methodology}

\subsection{Study Objectives}

We designed a controlled, mixed-design user study to empirically evaluate the efficacy of the SPHERE preference memory module. The primary objective is to validate that historically inferred spatial interactions can function as robust, reusable constraint signals for zero-shot scene generation. We hypothesize that the adaptive framework will significantly reduce the required corrective edit operations compared to a non-adaptive baseline, concurrently enhancing the user's perceived alignment of the synthesized layout with their internal spatial logic. The study content and procedures were approved by Institutional Review Board in advance.

\subsection{Experimental Procedure}

\subsubsection{Participants}

We recruited 42 participants ($N=42$; 23 female, 19 male) from a local university and the surrounding community, with ages ranging from 20 to 40 years ($M=27.3$, $SD=5.0$). 
All participants had normal or corrected-to-normal vision and no susceptibility to visually induced motion sickness (VIMS). 
To evaluate the system's utility for novice creators, individuals with professional expertise in 3D modeling or spatial design were excluded. Regarding prior experience, 52\% of participants had used VR once or twice, while 48\% reported more frequent exposure. Self-reported familiarity with VR controls was distributed across slightly (40\%), moderately (29\%), and very familiar (31\%) levels.

\subsubsection{Apparatus and Interface Training}

The immersive environment was rendered utilizing a Meta Quest 3 head-mounted display (HMD)\footnote{\url{https://www.meta.com/quest/quest-3/}}, with 6-DoF spatial interactions mediated via handheld controllers. Participants interacted with the spatial canvas through a multimodal fusion of controller-grounded manipulation and speech-actuated editing. Prior to the experimental trials, participants underwent a standardized onboarding protocol within a controlled, neutral \textit{Training Scene (Scene S)}. An experimenter demonstrated essential navigational mechanics alongside three primary editing modalities: (i) spatial translation, (ii) rotational manipulation, and (iii) voice-command attribute modification (e.g., semantic color replacement). Participants were required to demonstrate operational proficiency in these primitives before advancing to the evaluated tasks.

\subsubsection{Experimental Design}

We implemented a $2 \times 3$ mixed factorial experimental design. The between-subjects independent variable was the activation state of the \textit{preference memory module}. Participants were randomized into an \textbf{Adaptive condition} (where latent preferences extracted from the initial scene directly parameterized subsequent generations) or a \textbf{Baseline condition} (where sequential scenes relied exclusively on static, default synthesis heuristics without historical context). 

The within-subjects variable was the \textit{spatial constraint environment} (Rounds 1–3). Participants completed three consecutive generation-and-editing tasks within the same semantic category (living room), while the underlying room geometry was systematically perturbed across rounds to evaluate constraint transferability:

\textbf{Scene 1 (Standard Baseline):}
An isotropic, near-cubic spatial volume (e.g., $W=60$, $D=60$, $H=40$) is utilized to capture baseline behavioral tendencies and instantiate the initial interaction history under the assigned preference profile.

\textbf{Scene 2 (Scale Generalization):}
Introduced a severe volumetric scaling perturbation—either a contraction (e.g., $W=30$, $D=30$, $H=30$) or an expansion (e.g., $W=120$, $D=120$, $H=60$) relative to Scene 1. This condition isolates whether the system's learned constraints are rigidly overfitted to absolute coordinate frames or successfully generalize as relative topological structures and density distributions.

\textbf{Scene 3 (Topological Consistency):}
Introduced an asymmetric aspect ratio constraint (e.g., $W=60$, $D=20$, $H=40$). This induced severe geometric stress, designed to evaluate the system's capacity to resolve spatial conflicts, forcing trade-offs between strict metric adherence and the preservation of core relational topology.

\subsubsection{Target Preference Profiles}

To rigorously evaluate the system's capacity to capture and re-apply spatial logic while mitigating the confounding variance of evolving human aesthetic whims, we conducted a controlled evaluation using pre-defined target preference profiles. Rather than measuring unconstrained, in-the-wild personalization—which is inherently subjective, noisy, and difficult to standardize—this controlled setup provides an objective, reproducible baseline. It allows us to quantitatively isolate and evaluate how effectively SPHERE extracts hierarchical spatial rules and transfers them across varying geometric constraints. Specifically, participants were assigned to explicitly manifest one of three predefined profiles, which dictate deterministic constraints across a hierarchical spatial schema: \textit{Object-Level} (semantic attributes), \textit{Intra-Area Level} (local relational topology), and \textit{Inter-Area Level} (global spatial flow). Participants were instructed to iteratively refine each generated scene until it achieved strict conformity with their assigned profile.


\textbf{Profile A (Functional Viewer):}
Optimizes for linear media consumption and spatial efficiency. Target constraints mandate cool-toned palettes and modern sectional seating (Object-Level); strict parallel alignment and centering between seating elements and the primary display interface (Intra-Area Level); and discrete, spatially delineated walkways to isolate the entertainment nucleus from secondary functions, such as dining (Inter-Area Level).

\textbf{Profile B (Social Connector):}
Optimizes for face-to-face social affordances and ambient warmth. Target constraints mandate warm ambient illumination and discrete armchair entities (Object-Level); centripetal, circular, or semi-circular seating arrangements to maximize mutual gaze (Intra-Area Level) ; and an anchored centralization of the primary conversational cluster, deliberately detaching furniture from boundary walls and corners (Inter-Area Level).

\textbf{Profile C (Spatial Minimalist):}
Optimizes for negative space and fluid circulation. Target constraints mandate a strict minimal asset count and monochromatic texturing (Object-Level) ; maximal spatial dispersion between instantiated entities to reduce local density (Intra-Area Level) ; and strict perimeter alignment of major volumetric objects to preserve a contiguous central void (Inter-Area Level).

\subsubsection{Experimental Procedure and Task Flow}

The experimental protocol required approximately 60–75 minutes per participant, structured into three phases:

\textbf{Phase 1 (Onboarding \& Training, $\sim$15 min):}
Acquisition of informed consent and demographic logging, followed by the mandatory familiarization phase within Scene S.

\textbf{Phase 2 (Generative Authoring, $\sim$40 min):}
Participants iteratively completed the three scene rounds: \textbf{Round 1 (Scene 1):} The system synthesized an initial standard layout. The participant executed edits until reaching profile conformity, logging the full interaction graph. In the Adaptive condition, this terminal state and interaction log were parsed to formulate the persistent preference memory. \textbf{Round 2 (Scene 2):} The system generated a layout under scale perturbation. The Adaptive framework conditioned this synthesis on the previously inferred constraints , whereas the Baseline framework utilized default agnostic heuristics. \textbf{Round 3 (Scene 3):} The system generated a layout under the extreme aspect ratio perturbation to assess topological stability under geometric stress.

\textbf{Phase 3 (Post-Hoc Evaluation, $\sim$15 min):}

Following the completion of Scenes 2 and 3, participants completed the NASA-TLX\cite{hart2006nasa} along with structured questionnaires designed to assess perceived spatial consistency, object style selection, functional organization, and system attribution, as described in Section~\ref{questionnaire}. Semi-structured interviews were additionally conducted. After Scene 3, participants further completed the System Usability Scale (SUS)\cite{lewis2018system} and a questionnaire measuring perceived system intelligence across four dimensions---\emph{Adaptability, Time Saving, Trust} and \emph{Ownership}---each rated on a 7-point Likert scale (1=Strongly Disagree, 7=Strongly Agree) This phase was designed to capture qualitative grounding regarding perceived system alignment, learning efficacy, and the experiential friction of edge-case failures.


\section{Metrics and Edit Categorization} 

To rigorously assess the system's efficacy, we conceptualize personalization not merely as an aggregate reduction in error metrics, but as a transparent, human-interpretable mechanism. Consequently, we designed an aspect-level attribution framework to systematically quantify which specific spatial properties users perceive as being actively influenced by their historical interactions. The evaluated dimensions encompass spatial distance, orientation and facing, functional structuring, macro-level zoning, and semantic object style selection.

To provide an objective behavioral corollary to subjective perception, all user-initiated edit operations were continuously logged and taxonomized into three distinct constraint-oriented categories: \textbf{Parametric Edits ($P$):} Corrections addressing quantitative spatial violations, specifically pertaining to micro-level distance, angular orientation, and axial facing. \textbf{Structural Edits ($S$):} Interventions rectifying topological, relational, or macro-zoning discrepancies within the scene graph. \textbf{Style Edits ($E$):} Modifications targeting semantic object identity, instantiation, or aesthetic attribute violations. For each synthesized scene, the aggregate corrective effort is formally computed as:
\begin{equation}
    E_{\text{total}} = P + S + E
\end{equation} 
Our evaluation analyzes both the absolute magnitude of $E_{\text{total}}$ and its proportional category composition to isolate the specific spatial dimensions where the adaptive framework exerts the highest influence.

To bridge the objective behavioral logs with subjective user perception, we formalized a deterministic mapping between the psychometric aspects and our edit taxonomy: Subjective spatial distance and orientation map directly to Parametric Edits ($P$). Subjective functional structuring and zoning map directly to Structural Edits ($S$). Subjective object style selection maps directly to Style Edits ($E$). This theoretical mapping enables us to empirically test the hypothesis that a higher perceived alignment in a specific spatial dimension directly correlates with a statistically significant reduction in its corresponding edit category.

\label{questionnaire}
Following the completion of the generalization tasks (Scenes 2 and 3), participants were administered a psychometric instrument utilizing a 7-point Likert scale ($1 = \text{Strongly Disagree}$, $7 = \text{Strongly Agree}$). The questionnaire was rigorously designed to measure four core constructs: \textbf{Spatial Consistency:} Perceived adherence to micro-level parametric constraints ($P$-related). \textbf{Object Style Selection:} Perceived adherence to semantic and aesthetic preferences ($E$-related). \textbf{Functional Organization:} Perceived adherence to macro-level topological relationships ($S$-related). \textbf{System Attribution:} The degree to which users explicitly recognized the personalization module's influence on the initial layout generation. The core analytical phase compares the Adaptive and Baseline conditions across both the objective edit frequencies and the subjective alignment scores, meticulously examining the correlational structure between the subjective alignment ratings and the category-specific objective edit distributions.

\begin{figure}
    \centering
    \includegraphics[width=1\columnwidth]{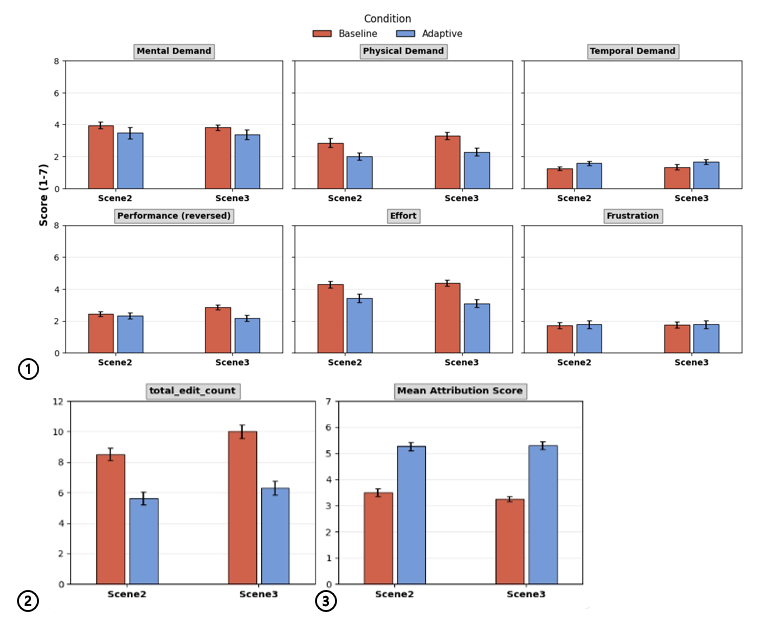}
    \caption{Comparison of subjective workload and editing outcomes across Scene 2 and Scene 3. (1) NASA-TLX subscale ratings for the Baseline and Adaptive conditions. (2) Total edit counts, showing fewer corrective edits in the Adaptive condition. (3) Mean attribution scores, indicating stronger perceived personalization in the Adaptive condition.}
    \label{fig:editcount&attribution}
\end{figure}

\section{Results}
We analyzed the results from the user study (N=42) to evaluate SPHERE's impact on editing effort and perceived personalization.

\begin{figure}[t]
    \centering
    \includegraphics[width=8.5cm]{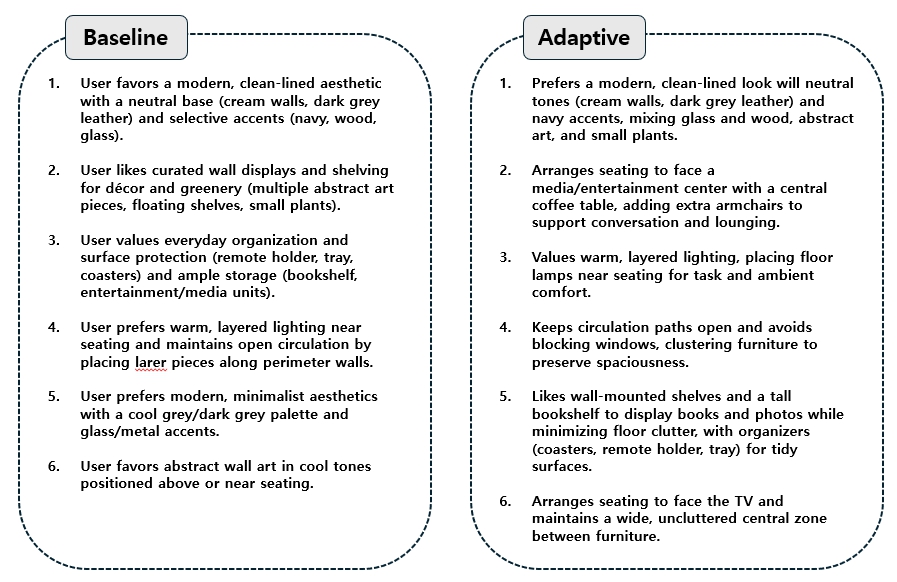}
    \caption{Example comparison of inferred preference constraints generated by the Baseline and Adaptive conditions for the same session. The Adaptive condition captures richer spatial organization and profile-consistent layout reasoning than the Baseline.}
    \label{fig:preferencecomparison}
\end{figure}

\subsection{Editing Effort}
As shown in Figure~\ref{fig:editcount&attribution}(2), the average total edit count in the Baseline and Adaptive conditions was 8.52 and 5.62, respectively. We used Welch's independent-samples $t$-test\cite{welch1938significance}  to examine the difference between the two conditions, and the result was statistically significant ($t = 4.86$, $p < .001$). This indicates that participants in the Adaptive condition required substantially fewer corrective edits than those in the Baseline condition, suggesting that the adaptive preference-based framework effectively reduced editing effort.

\subsection{Constraint-Level Perceived Personalization}
Overall, across Scene 2 and Scene 3, Table~\ref{tab:correlation_results} suggests that editing behavior was selectively aligned with perceived quality depending on the dimension and room characteristics. In both scenes, style satisfaction consistently showed a significant negative correlation with style-related edit counts, indicating that when participants perceived higher stylistic quality, they tended to make fewer corresponding edits. However, the lack of significant correlations for $P$ may be partly explained by the limitations of the system’s pointing-based editing control; because object manipulation was not always precise, participants may have repeatedly adjusted placements, inflating spatial edit counts regardless of their actual spatial satisfaction. Furthermore, in Scene 3, the narrow, corridor-like layout likely constrained object arrangement, making it more difficult to maintain stable and coherent functional groupings. As a result, functional quality may have been less consistently reflected in edit frequency compared to Scene 2. In Figure~\ref{fig:editcount&attribution}(3), participants in the Adaptive condition reported significantly higher attribution scores than those in the Baseline condition in both Scene 2 and Scene 3. In Scene 2, the mean attribution scores were 3.51 for the Baseline condition and 5.27 for the Adaptive condition, with a statistically significant difference between conditions ($t=-8.20$, $p<.001$). A similar pattern was observed in Scene 3, where the mean attribution scores were 3.27 for the Baseline condition and 5.30 for the Adaptive condition, again showing a significant difference ($t=-11.5$, $p<.001$). These results suggest that participants more strongly recognized the influence of the personalization module in the Adaptive condition across both generalization scenes. 

\subsection{Workload}
The mean NASA-TLX scores for the Baseline and Adaptive conditions were 3.10 and 2.89, respectively(see Figure~\ref{fig:editcount&attribution}(1)). While the Adaptive condition showed a slight advantage in overall workload, the difference was not statistically significant (Welch's $t(39.3)=1.349$, $p=.185$). One notable difference was that participants in the Adaptive condition reported significantly lower physical demand and effort than those in the Baseline condition. This suggests that, although the Adaptive system did not significantly reduce overall workload, it may have made the task feel less physically demanding and less effortful. Overall, we did not observe any meaningful increase in perceived task load.

\begin{figure}
    \centering
    \includegraphics[width=1\columnwidth]{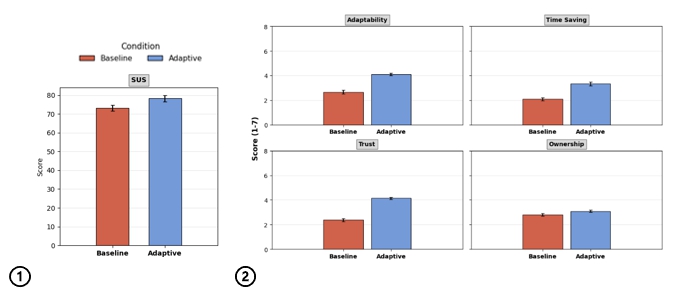}
    \caption{SUS and System's Perceived Intelligence results. The Adaptive condition received higher usability ratings than the Baseline and was also rated significantly higher in adaptability, time saving, and trust.}
    \label{fig:survey3}
\end{figure}

\subsection{System Usability}
As shown in Figure~\ref{fig:survey3}, the mean SUS scores for the Baseline and Adaptive conditions were 73.2 and 78.3, respectively. We used an independent-samples $t$-test to examine the difference between the two conditions, and the result was statistically significant ($t(40)=-2.20$, $p=.034$). This indicates that participants perceived the Adaptive condition as significantly more usable than the Baseline condition. Consistent with this result, participants in the Adaptive condition also reported significantly higher ratings for adaptability, time saving, and trust. Together, these findings suggest that the adaptive preference-based framework improved perceived usability and strengthened users' confidence in the system, particularly in terms of adaptability, efficiency, and trustworthiness.

\begin{table}[htbp]
\centering
\small
\setlength{\tabcolsep}{4pt} 
\caption{Correlation between perceived alignment scores and edit counts across Scene 2 and Scene 3. All correlations were computed with $df=40$.}
\label{tab:correlation_results}
\begin{tabular}{llcc}
\toprule
Scene & Variable Pair & Pearson's $r$ & $p$ \\
\midrule
\multirow{3}{*}{Scene 2} 
& Mean Spatial Score -- P\_count    & -0.082 & .605 \\
& Mean Style Score -- E\_count      & -0.424 & .005 \\
& Mean Functional Score -- S\_count & -0.334 & .031 \\
\midrule
\multirow{3}{*}{Scene 3}
& Mean Spatial Score -- P\_count    & -0.278 & .075 \\
& Mean Style Score -- E\_count      & -0.644 & $< .001$ \\
& Mean Functional Score -- S\_count & -0.218 & .165 \\
\bottomrule
\end{tabular}
\end{table}

\subsection{Offline Ablation Study}
\subsubsection{Experimental Setup}
Before evaluating the user experience, we conducted an offline ablation study to validate the architectural contribution of the Context Extraction Module. Specifically, we examined whether explicitly modeling both Local Context, which captures functional object clusters, and Global Context, which represents room-scale connectivity and layout structure, enables the system to infer deeper and more profile-consistent spatial preferences than relying on raw scene data alone. We used the dataset collected from the main user study and extracted the interaction logs and final scene JSONs from Round 1 (Scene 1). The Preference Inference Module was then re-run under two conditions. Condition A (Ours) used the full pipeline, in which raw scene data were augmented with derived Local Context obtained through DBSCAN-based clustering\cite{ester1996density} and Global Context represented as connectivity graphs prior to LLM inference. Condition B (Baseline) disabled the Context Extraction Module, requiring the LLM to infer preferences directly from raw interaction logs and absolute object coordinates without any intermediate contextual abstraction. This design isolates the role of contextual scene abstraction in preference inference while controlling for the downstream language model and the source behavioral data.

\subsubsection{Metrics}
We evaluated the inferred constraints along two dimensions: semantic alignment and structural diversity.

\textbf{Profile Alignment Rate}
To assess whether the extracted constraints reflected the assigned target preference profile, we employed a GPT-4-based evaluator\cite{achiam2023gpt}. For each session, the evaluator was provided with the profile description and the list of inferred constraints, and each constraint was classified into one of three categories: Aligned, if it supported the target profile; Neutral, if it was valid but not specifically informative of the target profile; and Misaligned, if it contradicted the target profile. We then computed the Profile Alignment Rate as the proportion of aligned constraints among all extracted constraints.

\textbf{Spatial Level Distribution}
Prior studies\cite{wu2024gpt, ccelen2024design, liu2023g} have shown that GPT-4-based evaluators can provide human-aligned judgments for subjective assessment of 3D content, supporting their use as an efficient and scalable evaluation method. To evaluate the granularity of the extracted preference knowledge, we categorized aligned constraints into three abstraction levels: Object-Level, Intra-Area Level, and Inter-Area Level. Object-Level constraints describe isolated object attributes such as color, category, or quantity. Intra-Area constraints describe topological relationships within a functional group, such as facing direction, distance, or surrounding arrangements. Inter-Area constraints describe the global placement of groups relative to the room boundaries or other groups.

This categorization was performed by two human raters (inter-rater reliability $\kappa > 0.560$). In cases where only one rater labeled a constraint as object-level, disagreements were resolved through adjudication, and a final consensus label was assigned.

\subsubsection{Quantitative Results}

Table~\ref{tab:combined_ablation_results} summarizes the Profile Alignment Rate for each profile and the overall average across conditions. Overall, Condition A achieved a higher alignment rate than Condition B (0.511 vs. 0.422), indicating that contextual abstraction improved the semantic consistency of inferred preferences with respect to the assigned target profiles. At the profile level, Condition A outperformed Condition B for Profile A (0.600 vs. 0.333) and Profile C (0.467 vs. 0.400), while Condition B showed a slightly higher alignment rate for Profile B (0.533 vs. 0.467). Despite this exception, the overall trend suggests that incorporating Local and Global Context generally helps the model discover profile-relevant spatial intentions more accurately. Table~\ref{tab:combined_ablation_results} reports the spatial level distribution of extracted constraints. The difference between conditions was particularly pronounced in terms of structured diversity. When examining the proportion of Object-Level constraints among all extracted constraints, Condition A yielded a ratio of 0.35, whereas Condition B yielded a substantially higher ratio of 0.75. This indicates that the baseline was heavily skewed toward simple attribute or object-presence level descriptions, while the full-context pipeline produced a more balanced set of constraints that included richer intra-area and inter-area spatial reasoning. Although the annotations showed only moderate agreement, such variability is expected for inherently subjective criteria and does not necessarily invalidate the evaluation. Nevertheless, the results should be interpreted cautiously, and future work should incorporate broader human validation to improve reliability. Overall, contextual modeling improves semantic alignment and structural diversity by capturing the relational and layout-level patterns central to spatial personalization.

\begin{table}[htbp]
\centering
\small
\setlength{\tabcolsep}{4pt} 
\caption{Profile alignment accuracy and proportion of object-level constraints (O / total) across conditions. For constraint categorization, evaluation was conducted by two human raters ($\kappa > 0.560$), and disagreements where only one annotator categorized a constraint as object-level were resolved through adjudication.}
\label{tab:combined_ablation_results}
\begin{tabular}{lccccc}
\toprule
 & \multicolumn{4}{c}{\textit{Profile Alignment Accuracy}} & \textit{Granularity} \\
\cmidrule(lr){2-5} \cmidrule(lr){6-6}
Condition & Profile A & Profile B & Profile C & Total & O / total \\
\midrule
Condition A & 0.600 & 0.467 & 0.467 & 0.511 & 0.35 \\
Condition B & 0.333 & 0.533 & 0.400 & 0.422 & 0.75 \\
\bottomrule
\end{tabular}
\end{table}

\subsubsection{Qualitative Results}

To illustrate the condition differences, We examined a Profile A sample session (Figure~\ref{fig:preferencecomparison}). The baseline primarily produced vague, generic constraints regarding atmosphere, style, and mere object placement, weakly grounded in interaction history and lacking the profile's underlying spatial logic. In contrast, our method generated both object-level preferences and higher-level constraints capturing spatial organization. Crucially, it consistently inferred functional rationales aligned with the target profile—such as maintaining open circulation or task comfort. This suggests that my full-context pipeline supports a structured, profile-consistent reasoning process, advancing from surface-level stylistic descriptions to spatially meaningful preference representations.

\section{Discussion}

The empirical evaluation of SPHERE demonstrates that extracting hierarchical spatial constraints from user interactions significantly enhances the adaptive generation of VR indoor scenes. By bridging the semantic gap between isolated generative instances, our framework not only reduces user effort but also establishes a more resilient and interpretable personalization pipeline.

\subsection{Mitigating Interaction Friction in Immersive Environments.}
A primary contribution of SPHERE is the marked reduction in the physical burden associated with immersive scene authoring. While the overall cognitive workload (NASA-TLX) did not show a statistically significant decrease, the significant reduction in physical demand and effort is a critical finding. Spatial manipulation in VR—particularly continuous locomotion, reaching, and precise object placement—is inherently fatiguing over extended sessions. As evidenced by the significant drop in total edit counts ($t = 4.86, p < .001$), the adaptive framework successfully offloads the heavy lifting of macro-level spatial organization to the system. By initializing scenes that already respect the user's underlying spatial logic (e.g., facing directions, circulation paths), users are freed to focus on micro-level aesthetic refinements rather than structural overhauls. This shift from physical construction to curatorial refinement is further reflected in the significantly higher usability (SUS) and adaptability ratings reported by participants in the Adaptive condition.

\subsection{The Necessity of Contextual Abstraction.}
The offline ablation study highlights a fundamental limitation of relying solely on raw LLM inference for spatial reasoning. When the Context Extraction Module was disabled (Condition B), the extracted constraints were overwhelmingly skewed toward object-level attributes (0.75 ratio), such as color and mere presence. The LLM struggled to deduce relational layouts from absolute coordinate data, often hallucinating vague stylistic justifications. Conversely, by explicitly modeling Local Context (functional clusters via DBSCAN) and Global Context (bounding volumes and geometric centroids), Condition A empowered the LLM to “see” the scene structurally. This abstraction proved vital for spatial resilience. When subjected to severe geometric perturbations in Scenes 2 and 3, constraints rooted in relative topology (e.g., maintaining an open central void) generalized successfully, whereas constraints tied to absolute coordinates would have failed catastrophically. Thus, effective spatial modeling must evolve from mere object presence (\textit{what}) to relational layout logic (\textit{how} and \textit{why}).

\subsection{Implications for Human-AI Co-Creation.} Beyond usability, SPHERE advances human-AI co-creation in immersive systems by translating behavioral traces into reusable spatial logic. This human-centered approach ensures generative models synthesize predictable, personalized outcomes rather than relying on generic priors. The significantly higher system attribution scores ($p < .001$) in the Adaptive condition indicate users actively perceive this collaborative loop. When the AI visibly applies learned constraints, it builds trust and perceived intelligence—allowing users to steer a responsive collaborator rather than fight against a stochastic ``black box.'' Although the baseline comparison evaluates the overall benefit of SPHERE, it does not fully isolate the contribution of each individual component. Future work should therefore include more fine-grained component-level baselines and user studies to quantify their respective effects.

\section{Limitations and Future Work.}
While SPHERE demonstrates strong adaptive capabilities, several limitations present opportunities for future research: \textbf{Control Imprecision:} As observed in Scene 3, the lack of a significant correlation between spatial satisfaction and parametric edit counts ($P$) was partially confounded by the limitations of pointing-based VR controls. Imprecise ray-casting often necessitated repeated micro-adjustments, artificially inflating edit frequencies. Future iterations should explore intention-aware snapping mechanisms or multi-modal disambiguation to filter out noisy interaction data. 

\subsection{Extreme Geometric Constraints} 
As observed in Scene 3, highly restrictive or asymmetric topologies (e.g., narrow corridor-like layouts) impose severe spatial stress that challenges rigid functional groupings. Currently, SPHERE lacks an explicit constraint relaxation policy, meaning conflicting user preferences cannot always be gracefully resolved under severe volumetric compression, occasionally resulting in visual clutter or layout degradation. To address this, future iterations should incorporate a hierarchical constraint relaxation mechanism that dynamically negotiates spatial trade-offs during optimization. Rather than treating all retrieved preferences ($C_{\text{pref}}$) with equal rigidity, the generative engine should prioritize global topological connectivity (e.g., preserving unobstructed circulation paths and clearances) as non-negotiable hard constraints. Conversely, micro-level parametric alignments and exact Euclidean distances should function as soft constraints with dynamically expanding error tolerances. Furthermore, under extreme compression where a full functional group is physically unfeasible, an intelligent asset-pruning and semantic-substitution layer (e.g., automatically downgrading an expansive coffee table to a compact side table or reducing secondary seating) could be introduced.


\subsection{Temporal Dynamics of Preference} SPHERE deliberately updates its preference memory post-session to prevent visual churn, bypass computational latency, and filter out transient trial-and-error manipulations. While this design effectively optimizes short-to-medium authoring sessions by capturing the user's finalized spatial intent, it inherently defers system adaptation until the subsequent generation cycle. For extended, open-ended authoring workflows where user design goals might evolve over several hours, waiting until session termination may limit real-time responsiveness. A promising avenue for future research is exploring real-time, intra-session reinforcement that incrementally adapts generation heuristics ``on the fly'' without interrupting the user's creative flow—for instance, by deploying asynchronous background agents to silently update affordance weights, or by applying live topological refinements exclusively to out-of-view or uncurated room zones.


\subsection{Controlled Profile Adaptation vs. In-the-Wild Personalization} A primary methodological consideration of our evaluation is the reliance on pre-defined target profiles. While this controlled design was essential to establish an objective, reproducible baseline and isolate the system's spatial reasoning capabilities from pointing-based interaction noise, it inherently evaluates ``profile-aligned adaptation'' rather than unconstrained, long-term personalization. In real-world immersive co-creation, human preferences are rarely static or internally consistent; they are implicit, context-dependent, temporally evolving, and occasionally contradictory. While SPHERE successfully demonstrates the capacity to capture and transfer deterministic spatial rules within a structured session, effective in-the-wild personalization must account for spontaneous aesthetic drift and shifting user intent over extended periods. A critical challenge for future work will be developing longitudinal preference-learning architectures capable of filtering out behavioral noise, resolving contradictory user edits over weeks or months, and dynamically balancing historical preference persistence against spontaneous creative exploration.

\section{Conclusion}

We introduce SPHERE, an adaptive VR framework that transforms repetitive immersive authoring into a continuous learning process by modeling temporal preference persistence. By abstracting multimodal interactions into hierarchical spatial constraints, SPHERE captures the relational logic of user intent. Our evaluations confirm that this contextual abstraction significantly reduces corrective edits and physical demand, yielding geometrically resilient scenes that accurately reflect demonstrated spatial profiles. Ultimately, SPHERE provides a structured foundation for controlled spatial adaptation in immersive human-AI co-creation. 

\acknowledgments{%
This work was supported by the Ministry of Science and ICT (MSIT), Korea, through the Institute for Information \& Communications Technology Planning \& Evaluation (IITP), under the Graduate School of Virtual Convergence Support Program (IITP-2026-RS-2023-00254129), the AI Star Fellowship Support Program at Sungkyunkwan University (No. RS-2025-25442569), and the Development of XR Content Agent Technology Based on Emotion and Sensibility Reasoning at Sungkyunkwan University (No. RS-2026-25520944). This work was also supported by the Sports and Tourism R\&D Program through the Korea Creative Content Agency (KOCCA), funded by the Ministry of Culture, Sports and Tourism in 2024, under the project titled “Development of game-based digital therapeutics technology for adolescent mental health (psychological and behavioral control) management” (grant number: RS-2024-00344893). 
}

\bibliographystyle{abbrv-doi-hyperref}

\bibliography{template}

\end{document}